\pdfoutput=1
\documentclass[11pt]{article}

\usepackage{EMNLP2023}

\usepackage{times}
\usepackage{latexsym}

\usepackage[T1]{fontenc}
\usepackage[utf8]{inputenc}

\usepackage{microtype}

\usepackage{inconsolata}
\usepackage{graphicx}
\usepackage{xcolor}
\usepackage{amsmath}
\usepackage[ruled]{algorithm2e}
\usepackage{booktabs}
\usepackage{colortbl}
\usepackage{tcolorbox}
\usepackage{enumitem}
\usepackage{roboto}  % If you want to use Roboto
\usepackage{lipsum}  % Optional, for generating text
\usepackage{tabularx}  % For tables that stretch to a given width
\usepackage{float}

\title{Digital Twin Ecosystem for Oncology Clinical Operations}

\author{
    \small Himanshu Pandey \\
    \texttt{\small himanshu@risalabs.ai}
    \And
    \small Akhil Amod \\
    \texttt{\small akhil@risalabs.ai}
    \And
    \small Shivang \\ 
    \texttt{\small shivang@risalabs.ai}
    \And
    \small Kshitij Jaggi \\
    \texttt{\small kj@risalabs.ai}
    \AND
    \small Dr. Ruchi Garg \\ 
    \texttt{\small docruchi@risalabs.ai}
    \And
    \small Abheet Jain \\ 
    \texttt{\small abheet@risalabs.ai}
    \And
    \small Vinayak Tantia \\ 
    \texttt{\small vinayak@risalabs.ai}
    \AND
    \small RISA Labs Inc.
}

\begin{document}

\maketitle
\begin{abstract}
Artificial Intelligence (AI) and Large Language Models (LLMs) hold significant promise in revolutionizing healthcare, especially in clinical applications. Simultaneously, Digital Twin technology, which models and simulates complex systems, has gained traction in enhancing patient care. However, despite the advances in experimental clinical settings, the potential of AI and digital twins to streamline clinical operations remains largely untapped. This paper introduces a novel digital twin framework specifically designed to enhance oncology clinical operations. We propose the integration of multiple specialized digital twins, such as the Medical Necessity Twin, Care Navigator Twin, and Clinical History Twin, to enhance workflow efficiency and personalize care for each patient based on their unique data. Furthermore, by synthesizing multiple data sources and aligning them with the National Comprehensive Cancer Network (NCCN) guidelines, we create a dynamic Cancer Care Path—a continuously evolving knowledge base that enables these digital twins to provide precise, tailored clinical recommendations.
\end{abstract}

\section{Introduction}

Delivering optimal oncology care presents several challenges that complicate decision-making for healthcare providers. Although Electronic Health Records (EHRs) contain extensive patient data, there is often no user-friendly tool to easily access a patient’s complete treatment and medication history. This lack of accessibility makes it difficult for providers to review past treatments and make informed decisions about future care. Additionally, aligning a patient's specific condition with the National Comprehensive Cancer Network (NCCN) guidelines poses another challenge. While the guidelines provide evidence-based recommendations, they must be personalized to each patient’s unique circumstances, such as their diagnosis, treatment history, and underlying conditions. Furthermore, even after identifying the appropriate treatment, it must adhere to payer-specific Prior Authorization (PA) criteria, adding administrative complexity and potentially delaying care.

Artificial Intelligence (AI) and Large Language Models (LLMs) have greatly impacted healthcare by enabling faster, more precise care across various applications, including surgery, medical consultations, administrative workflows, treatment planning, diagnosis, and more \cite{vaananen2021ai}. A notable advancement in this field is the development of digital twin technology, which facilitates personalized treatment simulations and predictive modeling, directly impacting patient-specific outcomes. The concept of a Digital Twin (DT) is well-established and refers to a dynamic, real-time digital counterpart of a physical object or system, with continuous data exchange and interaction—distinguishing it from a static \textit{digital model} that lacks real-time data integration between the digital and physical systems \cite{kritzinger2018digital}. Advances in AI and LLMs have significantly accelerated the adoption of digital twins in healthcare \cite{9255249, makarov2024large}. In this context, digital twins can model human organs, tissues, and cells, constantly updating with new data to predict future states. Moreover, precision medicine harnesses these technologies to tailor disease treatment and prevention based on individual genetic, environmental, and lifestyle factors, leading to the creation of virtual human models that enhance clinicians' ability to understand, diagnose, and treat diseases effectively \cite{sun2023digital}.

While digital twin technology has revolutionized clinical applications in various medical specialties, its potential in enhancing healthcare operations remains largely under-explored. Digital twins can greatly assist in care coordination by helping providers create tailored care plans based on individual conditions. For instance, a coordinator digital twin could streamline communication between stakeholders, such as laboratories and healthcare providers, ensuring seamless information flow throughout the cancer care journey. Building upon the foundation of digital twins, agent-based models integrate AI and LLMs to create sophisticated multi-agent systems (MAS) that can simulate complex interactions within healthcare operations. These agent-based digital twins serve as digital counterparts for various assets in healthcare operations, enhancing accuracy and efficiency. Digital twins can model not only physiological elements like heart functions but also complex processes, such as clinical decision support and workflows like prior authorization \cite{croatti2020integration}.

This paper examines the concept of agent-based digital twins and their role in building LLM-based multi-agent systems for oncology operations. The objectives of this paper are as follows:

\begin{itemize}
    \item Define the digital twin framework, highlighting various capabilities and demonstrating its use cases in oncology.
    \item Establish a \textit{Cancer Care Path} as a knowledge base to support digital twins in making clinical recommendations.
    \item Present a case study illustrating how different agents collaborate using various knowledge bases to streamline workflows.
\end{itemize}
\section{Related Work}

AI and LLMs have revolutionized healthcare in recent years. Models like ChatGPT \cite{openai2024gpt4technicalreport} have matured to a point where they can significantly influence clinical decision-making \cite{wojcik2023beyond}, enhancing the care provided by human healthcare professionals and improving patient outcomes \cite{javaid2023chatgpt}. LLM-based frameworks enable the creation of healthcare agents that integrate data sources, knowledge bases, and analytical models into their LLM-driven solutions \cite{abbasian2024conversationalhealthagentspersonalized}. Open-source models like LLaMA \cite{touvron2023llamaopenefficientfoundation} have demonstrated performance comparable to proprietary models, particularly in radiology examination questions \cite{adams2024llama}. LLMs show potential in generating accurate, concise, and comprehensive responses to patient queries with minimal risk of harm \cite{10.1001/jamanetworkopen.2024.4630}. Recent advancements in prompting techniques, such as Chain of Thought (CoT) \cite{NEURIPS2022_9d560961} and In-Context Learning (ICL) \cite{dong2022survey}, enable models to think step-by-step based on provided examples, improving both transparency and answer quality. Prompt optimization techniques \cite{wen2024hard, pryzant2023automaticpromptoptimizationgradient} have further boosted performance on specialized tasks. Additionally, self-consistency \cite{wang2023selfconsistency} has increased the reliability of LLM-generated outputs.

LLMs have made a significant impact on healthcare, delivering remarkable results across various applications. They have been utilized in generating discharge summaries \cite{ellershaw2024automated, williams2024evaluating} and improving care coordination \cite{nashwan2023enhancing, jung2024enhancing}. Many studies have focused on EHR-based applications, including information extraction \cite{gu2024scalable, cui2024multimodal, ahsan2023retrieving}, question-answering \cite{yan2024large}, and text-to-SQL parsing \cite{EHRSQL}. In oncology, LLMs have been extensively employed. Domain-specific LLM development has been explored, particularly for prostate cancer \cite{Tariq2024.03.15.24304362}, and they have been used for decision support in personalized oncology care \cite{10.1001/jamanetworkopen.2023.43689}. GPT-4 has demonstrated the ability to answer 85\% of examination-style multiple-choice questions from the American Society of Oncology \cite{10.1001/jamanetworkopen.2024.17641}. LLMs have been applied to breast cancer report generation \cite{naik2023synchronous} and to extract clinical information from radiology reports, such as breast ultrasounds \cite{chen2024burextractllamallmclinicalconcept} and lung lesions from medical images \cite{li2024automatedclinicaldataextraction}. Additionally, LLMs have been used to interpret ESMO\footnote{Refer: https://www.esmo.org/guidelines} and NCCN guidelines for lung cancer \cite{doi:10.1200/JCO.2024.42.16suppl.e13637}, as well as to generate clinical guidance trees for decision-making \cite{li2023meddmllmexecutableclinicalguidancetree}.

Digital Twins have been applied in oncology for predicting cancer progression and detecting metastases \cite{Batch2022, kim2022machine}, as well as in cardiology, where they simulate heart functions to optimize treatments through predictive modeling \cite{corral2020digital}. In neurology, digital twins assist in therapy planning and patient monitoring for conditions such as Alzheimer's and epilepsy \cite{10469438}. AI and Digital Twins also contribute to fields like stress management \cite{10579038}, nephrology \cite{Miao2023}, and autism care \cite{Iannone2024}, illustrating their impact across medical disciplines.

\begin{figure*}
    \centering
    \includegraphics[width=\textwidth]{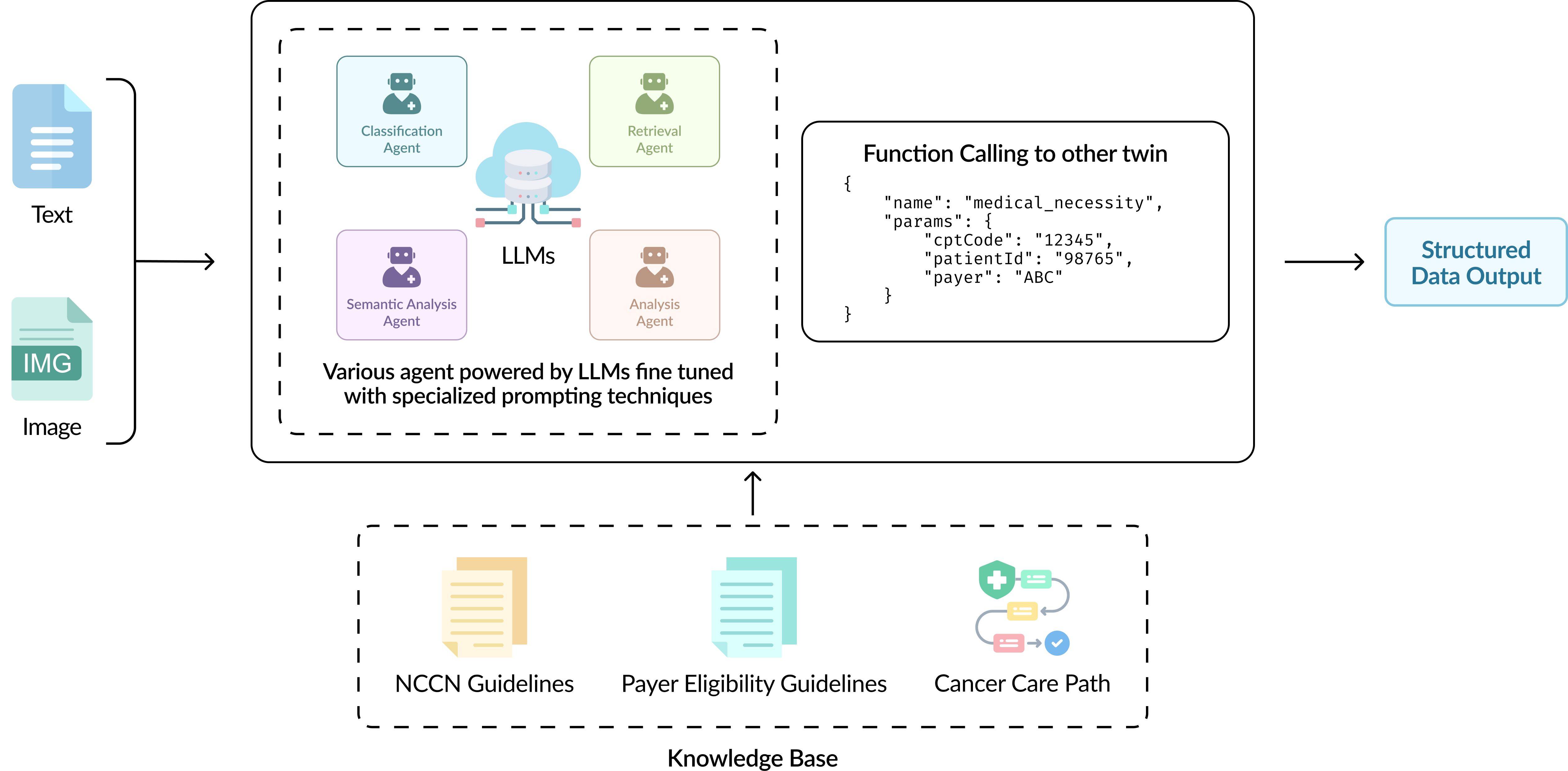}
    \caption{An illustration of the Digital Twin Framework highlighting its various capabilities and their integration with the knowledge base. }
    \label{fig:twin}
\end{figure*}

\section{Methodology}

Before delving into the practical applications, it is essential to introduce two key concepts that form the foundation of our approach. First is the \textit{Cancer Care Path}, which outlines the ideal sequence of care a patient should follow throughout their oncology journey. This serves as the benchmark for optimal patient care. Second, is the \textit{Digital Twin framework}, which defines the structure and functionality of our digital twins. It outlines the capabilities and inputs of each digital twin, as well as their interactions and collaboration with other agents to achieve desired outcomes. These principles collectively inform the design and implementation of digital twins in oncology care.

\subsection{Cancer Care Path}

\textit{Cancer Care Path} represents a comprehensive, structured approach of managing patient's journey, right from cancer diagnosis through treatment and beyond. It is designed to act as a knowledge graph or decision tree that outlines the ideal flow of care for a cancer patient. This path incorporates all the critical stages, starting from the early detection of symptoms, to the diagnostic tests that can be prescribed, and further to the available treatment options and medications. Each step or entity within this path is represented as an object in itself, having its own guidelines that dictate when and how it should be used. These guidelines ensure that decision-making is streamlined, reducing variability in care while improving patient outcomes by adhering to best practices. For example, an entity within the care path may represent a specific diagnostic test or treatment modality, and it comes with associated guidelines that help determine whether it is appropriate for a given patient. An agent, interacting with this path, can check whether all clinical criteria for proceeding to the next step are fulfilled. This system facilitates efficient, evidence-based care management.

NCCN Guidelines form the foundation of \textit{Cancer Care Path}. These evidence-based, research-backed, recommendations provide authoritative guidance on the treatment, management, and prevention of cancer. Developed by the \textit{National Comprehensive Cancer Network} (NCCN), these guidelines are widely regarded as gold standard in oncology for clinical decision-making. Furthermore, \textit{Cancer Care Path} has been developed in consultation with expert oncologists to ensure that it integrates practical clinical insights alongside NCCN's evidence-based recommendations. This comprehensive approach ensures that patients receive optimal care at every stage of their cancer journey, in alignment with the latest advancements in cancer treatment.

\subsection{Digital Twin Framework}

In our framework, a digital twin represents a virtual replica of a real-world entity or process within oncology care. This digital twin processes multiple input modalities—including text, audio, imaging, and structured medical data—and generates outputs based on predefined knowledge sources and embedded intelligence. These knowledge sources may include the cancer care path, NCCN guidelines, and payer guidelines for treatment acceptance. The digital twin’s functionality is driven by dynamically generated prompts, which are augmented through techniques like In-Context Learning (ICL) \cite{dong2022survey} and Chain-of-Thought (CoT) reasoning \cite{NEURIPS2022_9d560961}. In-Context Learning adapts to input data by retrieving examples from a vector database that stores relevant cases, utilizing Retrieval-Augmented Generation (RAG) \cite{NEURIPS2020_6b493230} to enhance decision-making processes. This retrieval ensures that the prompts remain contextually relevant to each individual patient case, thereby improving the twin’s ability to generate accurate and patient-specific insights.

Multimodal large language models (LLMs), potentially fine-tuned for oncology-specific tasks, process the dynamically generated prompts. By generating explicit reasoning logs, the digital twin’s outputs are made interpretable to healthcare professionals, offering insight into the rationale behind each prediction, recommendation, or simulation. To handle more complex tasks that go beyond the capabilities of large language models (LLMs), the digital twin ecosystem often integrates software engineering techniques. This integration transforms digital twins into sophisticated software programs that incorporate LLMs as core components, but extend their functionality through symbolic reasoning and algorithmic processes. For instance, these engineering techniques can be used to simulate the likelihood of a prior authorization request being approved by a payer, modeling not just probabilistic outcomes but also deterministic, rule-based logic such as payer-specific criteria.

In addition to its core functionality, the digital twin is capable of \textit{agent collaboration}, wherein it communicates and delegates tasks to other specialized agents via function calling. This distributed architecture ensures that complex, multi-step workflows are handled efficiently. For instance, if a twin requires verification of whether a treatment aligns with payer guidelines, it can delegate this task to another twin specifically designed to handle payer-related queries. This collaborative approach, as illustrated in Figure \ref{fig:ecosystem}, allows the ecosystem to manage interconnected processes in parallel, ensuring scalability and reducing the cognitive load on any single digital twin. By leveraging both multimodal data inputs and state-of-the-art machine learning techniques, the digital twin framework in oncology care provides an advanced, adaptive, and explainable system capable of supporting complex clinical decision-making.
\section{Twin Applications for Oncology}

\begin{figure*}
    \centering
    \includegraphics[width=\textwidth]{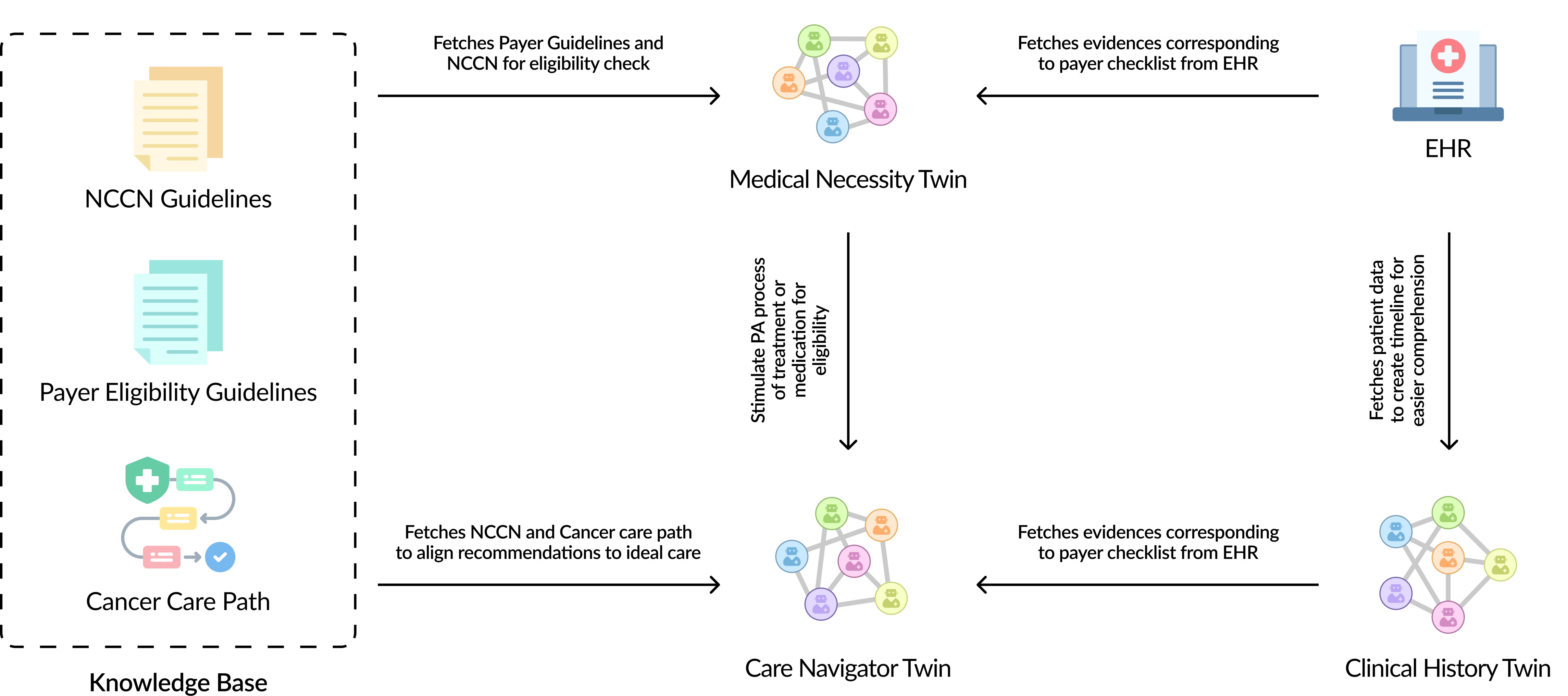}
    \caption{Different twins in Digital Twin ecosystem leveraging knowledge base and interacting with other twins for optimal efficiency}
    \label{fig:ecosystem}
\end{figure*}

The application of digital twins in oncology helps enhancing patient care by integrating advanced data analysis, clinical guidelines, and real-time decision support. Each twin is designed to handle specific aspects of the oncology care journey, from determining the medical necessity of treatments to visualizing a patient’s clinical history and guiding the overall care pathway. Together, these digital twins form a cohesive system that supports healthcare providers in making informed, evidence-based decisions at every stage of a patient’s treatment as depicted in Figure \ref{fig:ecosystem}.

\subsection{Medical Necessity Twin}

The \textit{Medical Necessity Twin} assists healthcare providers in determining the medical necessity of treatments and diagnostic tests in oncology care. By evaluating the patient's clinical data—such as symptoms, diagnostic tests, and prior treatments in conjunction with established guidelines such as the NCCN guidelines and payer-specific criteria, this twin ensures that proposed treatments are appropriate and justified based on the patient's medical history and condition.

\begin{figure}[H]
    \centering
    \includegraphics[width=\columnwidth]{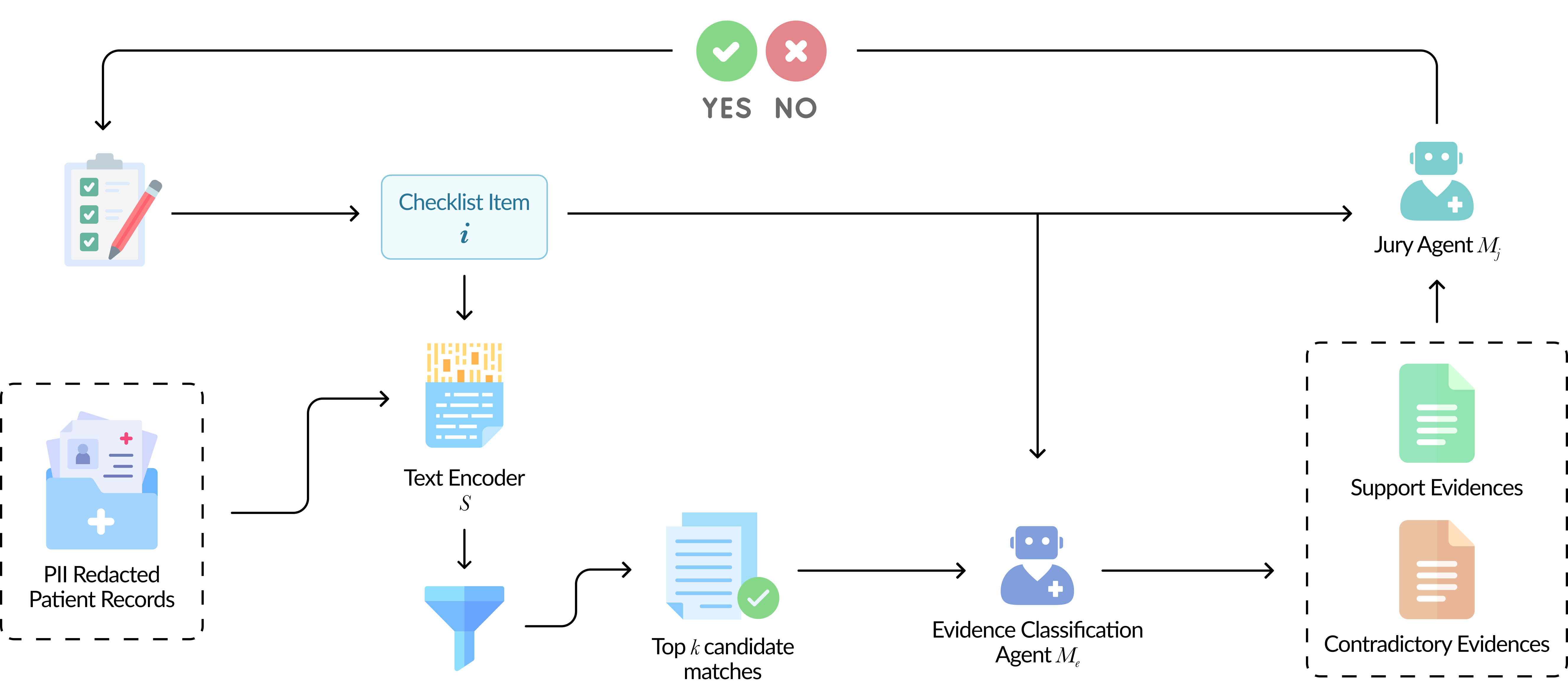}
    \caption{Illustrating the various agents that comprise the \textit{Medical Necessity Twin}, highlighting how they interact and address each item on the clinical guideline checklist.}
    \label{fig:medical_necesity_twin}
\end{figure}

As demonstrated in previous research, the Medical Necessity Twin achieves 86\% accuracy in automating the determination of medical necessity \cite{pandey-etal-2024-advancing}. Additionally, the Medical Necessity Twin supports other twins by providing real-time assessments of treatment eligibility. This ensures that any recommendations made by other twins meet the specific criteria required for payer approval, thereby improving the overall efficiency of patient care.

\subsection{Clinical History Twin}

The \textit{Clinical History Twin} is designed to visualize the complete timeline of a patient’s tests, treatments, and clinical interventions in a chronological format. By presenting this data in a clear, time-sequenced manner, this twin allows healthcare providers to quickly and efficiently interpret the progression of care. This visualization aids in identifying patterns, understanding the effectiveness of previous treatments, and determining the next steps in care. Furthermore, it allows healthcare professionals to spot gaps or delays in care and assess how well the treatment plan has been followed.

To implement the \textit{Clinical History Twin}, a combination of structured and unstructured data is extracted from the patient’s EHR. LLM-based agents are used to extract and convert key clinical events into structured formats identifying important events, such as treatment milestones or changes in a patient's condition, and organizing them within the timeline. Additionally, these agents map relationships between various clinical events—such as linking a diagnosis to subsequent treatments—thereby ensuring the timeline offers a comprehensive, interconnected view of the patient’s care history. Experiments to validate this approach are ongoing, and the results with a detailed architecture will be discussed in a future paper.

\subsection{Care Navigator Twin}

The \textit{Care Navigator Twin} plays a pivotal role in guiding a patient's oncology care journey by leveraging the cancer care path. By analyzing patient-specific data, including their EHR and clinical history, it predicts the next optimal steps in the care journey, ensuring that each decision follows the most current clinical guidelines. This involves identifying which diagnostic tests, treatments, or medications should be recommended at each stage of the patient's journey, based on both historical data and current best practices.

The Care Navigator Twin can anticipate potential challenges or deviations from the expected care path by comparing the ideal cancer care path with the patient’s actual care journey using input from the \textbf{\textit{Clinical History Twin}}. Additionally, its predictions are further strengthened by real-time data from the \textbf{\textit{Medical Necessity Twin}}, ensuring that treatment decisions are both clinically sound and meet payer-specific guidelines, increasing the likelihood of treatment approval.
\begin{figure*}
    \centering
    \includegraphics[width=\textwidth]{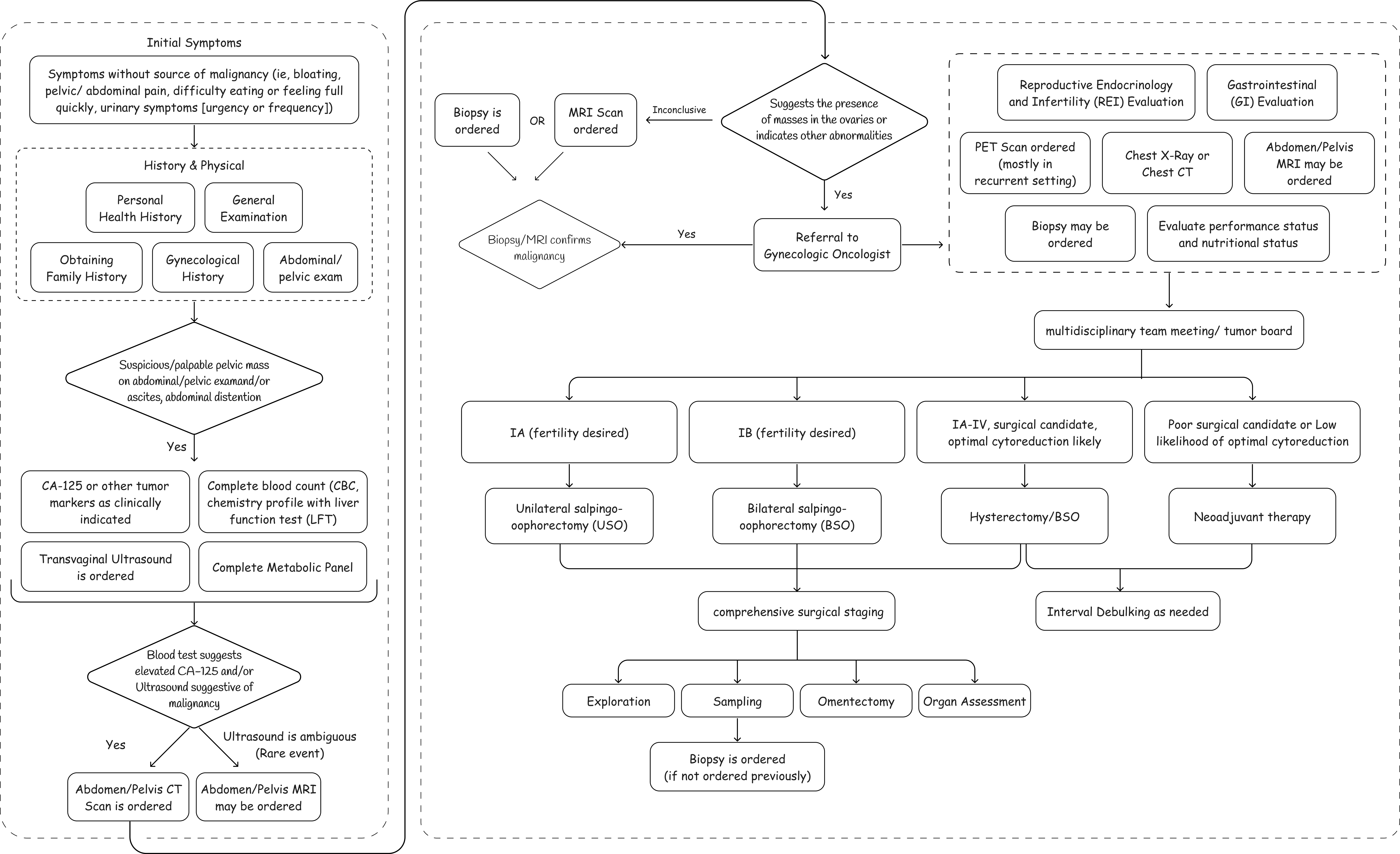}
    \caption{A visual representation of Cancer Care Graph built for Ovarian Cancer Diagnosis}
    \label{fig:diagnosis}
\end{figure*}

\section{Case Study: Ovarian Cancer Diagnosis}

We present a case study of ovarian cancer, demonstrating the use of the previously mentioned twins at various points in the care process. First, we constructed the cancer care graph using a variety of guidelines and online resources, particularly adhering to NCCN guidelines. We then illustrate how and where different agents interact throughout the patient's oncology care journey.

\subsection{Cancer Care Graph}

Using the ovarian cancer NCCN guidelines\footnote{Refer: https://www.nccn.org/guidelines/category\_1}, we extracted key diagnostic tests and tools employed in diagnosing ovarian cancer. Our focus is primarily on the initial symptoms and the diagnostic process that progresses to a clinical determination of ovarian cancer a hown in Figure \ref{fig:diagnosis}. Similar graphs can be developed for treatment and follow-up monitoring, and can be extended to other cancer types.

\subsection{Diagnosis Journey}

The journey begins when a patient presents to a primary care provider with symptoms such as bloating, pelvic or abdominal pain, difficulty eating or feeling full quickly, or urinary symptoms. If the provider suspects ovarian cancer after reviewing the patient’s history and conducting a physical exam, the \textbf{\textit{Care Navigator Agent}} suggests a series of preliminary tests based on NCCN guidelines. These tests typically include CA-125 or other tumor marker tests, a Complete Blood Count (CBC) with Liver Function Tests (LFT), and/or a transvaginal ultrasound. At this point, the \textbf{\textit{Medical Necessity Twin}} identifies the patient's payer from the EHR and retrieves the payer's guidelines for each recommended test. It then runs simulations to determine which of these tests are likely to be approved based on the patient’s medical condition and the clinical examinations already performed by the provider. A sample CA-125 guideline from Anthem\footnote{Refer: https://www.anthem.com/dam/medpolicies/abc/ active/guidelines/gl\_pw\_e002881.html} is shown in Figure \ref{fig:anthem}. The Medical Necessity Twin evaluates the relevant data, checks if the clinical guidelines are satisfied, and does the same for all recommended tests. The provider can then see, in real-time, which tests are eligible for recommendation based on the results provided by the Medical Necessity Twin.

Based on the preliminary test results, the Care Navigator Twin interprets the cancer care graph and recommends the next steps, which could include an abdominal/pelvic CT scan. It also recommends that the CT scan be conducted with oral and IV contrast, following NCCN guidelines. Once again, the \textbf{\textit{Medical Necessity Twin}} becomes active, identifies the correct CPT code for this test—74177 (CT of the abdomen and pelvis with contrast) versus 74176 (without contrast)—retrieves the relevant guidelines, and determines if the test will be approved.

\begin{figure}
\begin{tcolorbox}[
    arc=1mm, % Rounded corners
    colback=gray!10, % Light grey background
    colframe=gray!80, % Dark grey frame
    boxrule=0.5pt,
    fonttitle=\bfseries\scriptsize, % Smaller bold font for the title
    title=Tumor Marker Cancer Antigen 125 (CA-125) Testing,
    fontupper=\scriptsize\fontfamily{pcr}\selectfont, % Monospace font, smaller text
    left=3mm, % Reduces left padding
    right=3mm, % Reduces right padding
    top=2mm, % Reduces top padding (optional)
    bottom=2mm % Reduces bottom padding (optional)
]
\raggedright
\textbf{CA-125 testing is considered medically necessary for any of the following:}
\begin{enumerate}[leftmargin=*]
    \item Evaluation of a pelvic or abdominal mass in postmenopausal individuals; or
    \item Evaluation of a pelvic or abdominal mass suspicious for an epithelial ovarian cancer or other specified malignancy in premenopausal individuals; or
    \item Evaluation of an individual with signs or symptoms suggestive of ovarian cancer (for example, ascites, abdominal distention, bloating, pelvic/abdominal pain, difficulty eating or feeling full quickly, urinary symptoms);
\end{enumerate}
\end{tcolorbox}
\captionof{figure}{An example checklist from Anthem listing medical necessity conditions for CA-125 testing}
\label{fig:anthem} % Label for referencing this figure
\end{figure}

Similarly, for all other tests and treatments, as shown in Figure \ref{fig:ecosystem}, the Medical Necessity Twin fetches the appropriate guidelines from both the payer and NCCN sources. Based on simulations, it provides results to the provider. During each encounter with the provider, the \textbf{\textit{Clinical History Twin}} continues to accumulate data, offering insights from prior test results to both the provider and the Care Navigator Twin, thus enhancing the decision-making process.
\section{Conclusion}

In this paper, we introduced a framework for creating digital twins and demonstrated its use cases in oncology. Specifically, we presented three key twins: the Medical Necessity Twin, the Care Navigator Twin, and the Clinical History Twin. We showcased how these twins will collaborate to enhance the efficiency of care delivery. Additionally, we introduced the concept of the Cancer Care Path, which supports the twins in knowledge retrieval tasks. Combined with NCCN guidelines and payer-specific criteria, this forms a comprehensive knowledge base that the twins can leverage. This led to the creation of a collaborative ecosystem, where the twins operate within and are enriched by their environment. Each agent generates logs through chain-of-thought prompting, contributing to the explainability of the entire twin ecosystem. Finally, we presented a case study to illustrate the structure of the Cancer Care Path and how these agents work together to successfully navigate the care process for optimal outcomes in ovarian cancer.
\section{Future Work}

We have introduced an ecosystem that supports clinical decision-making and operations in oncology. While we defined a set of preliminary twins within this ecosystem, there is potential to expand the number of twins to support a broader range of operations. Each twin will adhere to the digital twin framework, enabling seamless collaboration and interaction within the ecosystem. The framework itself can also evolve to incorporate additional features that streamline collaboration and workflow management. One such enhancement could be the implementation of self-verification mechanisms, such as self-consistency or fact-checking, to minimize hallucination and ensure high-quality results.

Future twins could focus on tasks such as ambient note-taking, patient education, self-care best practices, and patient scheduling, each of which would benefit from the collaborative ecosystem. The knowledge base, which the twins rely on, must also become more sophisticated. It should not only include clinical guidelines but also practical insights, providing comprehensive and exhaustive knowledge tailored to specific use cases. Further work is needed to improve how knowledge is organized and presented to these agents, ensuring that the system produces accurate and reliable outputs.

While Chain of Thought (CoT) prompting enhances explainability at the agent level, there is a need for additional tools to provide transparency at the twin and ecosystem levels. This would help make the entire ecosystem more interpretable and trusted by the medical community, fostering broader acceptance of this technology in clinical settings.

\bibliography{anthology,custom}
\bibliographystyle{acl_natbib}

\appendix

% \section{Example Appendix}
% \label{sec:appendix}

% This is a section in the appendix.

\end{document}